\documentclass[a4paper,12pt]{article}
\usepackage{graphicx}
\usepackage{amssymb}
\usepackage{amsmath}
\usepackage[a4paper,portrait,twocolumn=false,includeheadfoot,mag=1000,dvips]{geometry}
\usepackage[title]{appendix}
\usepackage{caption}
\usepackage{float}
\usepackage{color}
\usepackage{ulem}
\usepackage{mathrsfs}
\usepackage{type1cm}
\usepackage{marvosym}
\usepackage{amssymb}
\usepackage{wasysym}
\usepackage{subfigure}
\usepackage{multirow}
\usepackage{booktabs}
\usepackage[utf8]{inputenc}

\usepackage{graphicx}
\usepackage{epstopdf}

\usepackage{makecell}

\usepackage{enumerate}
\usepackage{gensymb}

\makeatletter

\newcommand{\Rmnum}[1]{\expandafter\@slowromancap\romannumeral #1@}
\makeatother

\numberwithin{equation}{section}

\newcommand{\sihao}{\fontsize{15pt}{\baselineskip}\selectfont}
\newcommand{\xiaosihao}{\fontsize{14pt}{\baselineskip}\selectfont}

\begin{document}  \xiaosihao

\newtheorem{The}{Theorem}
\newtheorem{exam}{Example}
\newtheorem{lem}{Lemma}
\newtheorem{de}{Definition}
\newtheorem{prop}{Proposition}
\newtheorem{cor}{Corollary}
\newtheorem{hyp}{Hypothesis}
\newtheorem{rem}{Remark}
\pagestyle{plain}

\setlength{\baselineskip}{20pt}
\setlength{\parskip}{0.4\baselineskip}

\clearpage
\begin{center}
{\sihao \textbf{``Adversarial Stress Testing'' of Lifetime Distributions} \\
}

{\normalsize by}

{\normalsize Nozer D. Singpurwalla \\ }
{\normalsize The George Washington University\\}
%

{\normalsize January, 2020}


%

\end{center}

%
\renewcommand{\abstractname}{{\large Abstract}}
\begin{abstract}
{\normalsize
In this paper we put forward the viewpoint that the notion of stress testing financial institutions and engineered systems can also be made viable appropos the stress testing an individual's \textit{\textbf{strength of conviction}} in a probability distribution. The difference is interpretation and perspective. To make our case we consider a game theoretic setup entailing two players, an \textit{adversarial} $\mathscr{C}$, and an \textit{amicable} $\mathscr{M}$. The underlying metrics entail a de Finetti style 2-sided bet with asymmetric payoffs as a way to give meaning to lifetime distributions, an adversarial stress testing function, and a maximization of the expected utility of betting scores via the Kullback-Liebler discrimination.  }\\

\noindent {\normalsize \textbf{Keywords:} \textit{Cross Entropy, Discrimination Function,  Subjective Probability, \\ \indent \indent \indent \indent  Utility}.\\}
\end{abstract}
\addtocounter{section}{-1}
\newpage
\section{\large Preamble: What is ``Adversarial Stress Testing''?}

The term ``stress testing'' as used here is not to be interpreted in same vein as that used in banking and finance, though the intent of both is similar. Furthermore, ``stress testing'' is also not to be seen as another label for \textit{\textbf{accelerated life testing}} in reliability and biometry; the two contexts are different.

During the financial crisis of the early twenties, banks and financial institutions were subjected to what is known as a ``stress test''. Its aim was to assess the robustness of these institutions in withstanding disruptions, and an ability to provide their intended services for a specified timeframe. Stress tests are generally conducted by regulators, who as representatives of the public, are mandated to be adversarial. The purpose of this article is to extend the concept of a stress test of an institution to that of a probability distribution (or a survival function). The metric of discussion here is a lifetime, though the underlying idea need not be limited to such a metric.

Stress testing a probability distribution is not the same as an \textit{accelerated life test} done in reliability, or a biostatistician's \textit{dose-response experiment}. In these two scenarios, one assesses an item's capacity to endure a physical force by changing the conditions of the test via a systematic increase of the stress or the dose. Each stress (dose) level spawns its own lifetime distribution and the challenge is one of extrapolation based on several such distributions. By contrast, under an adversarial stress testing of a probability distribution, one assesses the \textit{strength of conviction} of the individual proposing the distribution; as such there is only one distribution under discussion. This is done by changing the conditions of an underlying 2-sided bet by increasing its risk levels.

To summarize, the duality between an accelerated test and an adversarial stress test can be encapsulated via the statement that in the former one assesses an item's physical strength to endure, whereas in the latter, it is an individual's strength of belief that gets scrutinized. Furthermore, in accelerated testing one encounters a family of lifetime distributions, and any adversarial element, even if present, is not treated explicitly. By contrast, under an adversarial stress test the focus of attention is the credibility of the specifier of the distribution that gets evaluated.

There are two other comments to this preamble. The first is that the stress test of a financial entity is very much in the same spirit as an accelerated test with binary outcomes. The second is that the adversarial stress test, to be proposed here, would be a conducive instrument for validating the survivability of one of a kind items.


\section{\large Adversarial Behavior and Subjective Probability}

With the advent of active consumerism, demanding certification, and aggressive litigation, the need for the intensive testing of items and algorithms, under an adversarial flavor has gained increased prominence. Adversaries are individuals (or a group of individuals) whose expected utilities differ. Differences in expected utilities occur because of differences in their assessed probabilities or their innate utilities, or both. In rare circumstances, the expected utilities of adversaries could agree, even though their probabilities and utilities do not. For probabilities to be different, it is axiomatic that their interpretation be subjective. Thus the notion of subjective probabilities seems almost mandatory for any version of an adversarial set up. Neither the relative frequency, nor the propensity interpretation of probability will be meaningful for every adversarial scenario. Whereas the existence of subjective probabilities has been established by the likes of Ramsey (1931) and Savage (1954), its operationalization by de Finetti (1974) as a 2-sided monetary bet makes its meaning explicit. We shall lean on this operationalization of probability, recognizing that in doing so it is not possible to separate one's probability from one's utility for money, because the two are entangled.

Within the realm of adversarial scenarios, there are two general classes worthy of distinction. The first is where the adversaries participate in an economic or strategic conflict, with the intent of annihiliating each other. Such scenarios are best addressed by game-theoretic methods, where the actions of one adversary occur as a surprise to the other [cf. von Neumann and Morgenstern (1944)]. The second adversarial context is the one entailing the exchange of goods by buying, selling, or the certification of an entity. Here there could be genuine differences of opinion between the adversaries about the underlying probabilities, see, for example, Lindley and Singpurwalla (1991), (1993). The overall goal of both adversaries is to do common good bearing in mind the premise that both need each other to achieve the good. An example is the certification of an aircraft or a piece of software wherein one member, say a manufacturer $\mathscr{M}$, seeks approval and acceptance of his/her product by an adversary $\mathscr{C}$, who could be a consumer (or a regulator). It is not the intent of $\mathscr{C}$ to annihilate $\mathscr{M}$, though it is possible that $\mathscr{C}$'s actions may eliminate $\mathscr{M}$ from future participation. Situation's of this type also occur in jurisprudence under courtroom settings. In the manufacturer-consumer scenario, there is technically speaking, at most one active adversary; the other player is generally amicable. Specifically, $\mathscr{C}$ can be adversarial to $\mathscr{M}$, but not vice-versa. Indeed, $\mathscr{C}$ may choose not to be adversarial at all. In this case, the scenario boils down to the classical case of acceptance sampling for quality control, typically addressed by the Fisher-Neyman-Pearson-Wald test of a hypothesis.

An adversarial scenario can also arise when there is a single $\mathscr{C}$ and multiple $\mathscr{M}$'s, the latter being adversarial to each other, with the possible goal of an $\mathscr{M}$ annihilating the other $\mathscr{M}$'s. Game theory enters the picture whenever the matter of annihilation comes in play, and in the context of several $\mathscr{M}$'s the issue of coalitions among several $\mathscr{M}$'s, each coalition endeavoring to annihilate the others, becomes germane.

In what follows, we restrict attention to the case of a single $\mathscr{C}$ and a single $\mathscr{M}$, first articulating the case of $\mathscr{C}$ not being adversarial to $\mathscr{M}$, and then the case when $\mathscr{C}$ is an active adversary, because $\mathscr{C}$'s probability distribution of a lifetime does not align with that declared by $\mathscr{M}$, or $\mathscr{C}$'s is required to be demanding of the viability of $\mathscr{M}$'s product.

\section{\large The Dispositions of a Passive $\mathscr{C}$ and an Amicable $\mathscr{M}$}

Consider an item whose lifetime $Y$ takes values $y\geq 0$. Suppose that $\bar{F}(y)=P(Y\geq y)$ is absolutely continuous, with probability density $f(y)=-d\bar{F}(y)/dy$. It is common for the survival function $\bar{F}(y)$ to be specified by $\mathscr{M}$, and this is what we shall assume. Of relevance to $\mathscr{C}$ are lifetimes that are greater than, or equal to $y^{*}>0$; $y^{*}$ is known as the \textit{mission time} or a ``threshold'', and its specification tends to be $\mathscr{C}$'s prerogative. The onus of accepting $\mathscr{M}$'s $\bar{F}(y^*)$, or challenging it, is also up to $\mathscr{C}$. Suppose that $\mathscr{C}$ has no interest declared in challenging $\mathscr{M}$'s $\bar{F}(y^*)$, other than to accept or to reject the item based on what $\bar{F}(y^*)$ is. How must $\mathscr{C}$ make tangible sense of what $\bar{F}(y^*)$ really means? In other words, what is the operational import to $\mathscr{C}$ of $\mathscr{M}$'s $\bar{F}(y^*)$? By operational import, we mean a system of bets between $\mathscr{C}$ and $\mathscr{M}$ entailing rewards and penalties.

de Finetti provided an operational interpretation of $\bar{F}(y^*)$ for $y=y^*$ in particular, and for that matter, any $y\geq 0$. His notion was that of a 2-sided monetary bet with a linear state-dependent utility. Whereas de Finetti's focus was not on adversarial considerations, his operational interpretation paves a path towards how one can expand his setup to an adversarial situation wherein $\mathscr{C}$'s survival function for $Y$, $\bar{G}(y)$ differs from the $\bar{F}(y)$ of $\mathscr{M}$.

To appreciate this, let us first consider de Finetti's interpretation of \\$\bar{F}(y^*) \overset{def}{=} p$, for $0< p< 1$. Here, the number $p$ implies that $\mathscr{M}$ is willing to stake $p$ on the table in exchange for a reward of +1 from $\mathscr{C}$, if $Y\geq y^*$, and is prepared to lose to $\mathscr{C}$ the $p$ staked should $Y < y^*$. The bet so placed is against $\mathscr{C}$ who stands to gain $p$ if $Y < y^*$ (i.e. if $\mathscr{M}$ fails requirements), and to lose 1 if $Y \geq y^*$ (i.e. $\mathscr{M}$ has met requirements). The 2-sidedness of $\mathscr{M}$'s bet also requires that $\mathscr{M}$ also stake (1-p) in exchange of 1 if $Y < y^*$, and lose the (1-p) staked if $Y \geq y^*$.

Under the above two bets, one for $Y \geq y^*$ and the other against it, the only action a passive $\mathscr{C}$ need take is to choose the side of the bet. In what follows we always assume that $\mathscr{C}$ chooses the first bet. Then for any choice $y^*$ of $Y$, $\mathscr{C}$'s \textit{payoff (or reward) function} $S(y)$, $y \geq 0$, is of the form shown in Figure 1. Assuming that $p$ and 1 are monetary units, and assuming that $\mathscr{C}$'s utility for money is linear, the $S(y)$ of Figure 1 is also, $\mathscr{C}$'s utility for realizing a lifetime $Y \geq y$.
\begin{figure}[!ht]
  \centering
  \includegraphics[scale=0.8]{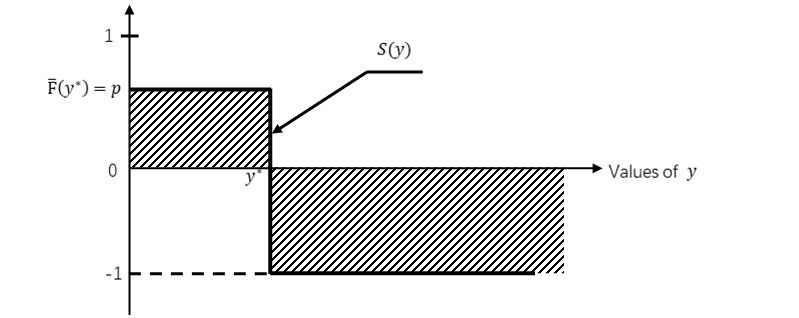}
  \caption{$\mathscr{C}$'s [$\mathscr{M}$'s] Payoff [Payback] Function $S(y)$}
\end{figure}

$\mathscr{C}$'s passive disposition to $\mathscr{M}$'s $\bar{F}(y)$ has resulted in a step function for $\mathscr{C}$'s payoff, wherein $\mathscr{C}$ receives a constant payoff of $+p$ for all unacceptable lifetimes, and also a constant payoff $-1$ for all acceptable lifetimes. In other words, $\mathscr{C}$'s utility for money is literally state independent. It was a feature like this -- among others -- that motivated Ramsey, Savage and others to develop a theory for the simultaneous axiomatization of probability and utility.

What must $\mathscr{C}$ do if the step-function payoff is not acceptable to $\mathscr{C}$, and/or if $\mathscr{C}$'s survival function for $Y$ $\bar{G}(y)$ is different from $\mathscr{M}$'s $\bar{F}(y)$? An obvious strategy would be for $\mathscr{C}$ to entice $\mathscr{M}$, to revise his/her $\bar{F}(y)$ to $\mathscr{C}$'s $\bar{G}(y)$, and repeat the 2-sided bet using $\bar{G}(y)$. However this approach merely translates the step-shaped payoff function; it does not change its overall character. An approach for changing the shape of the payoff to something more general is discussed later in Section 2.1. But before doing so, it is also instructive to bear in mind the shape of $\mathscr{M}$'s payoff function, when $\mathscr{C}$ chooses the first of de Finetti's 2-sided bet. Its general form is again a step-function that happens to be a 180$\degree$ rotation of $\mathscr{C}$'s payoff function about the horizontal axis; see Figure 2.
\begin{figure}[!ht]
  \centering
  \includegraphics[scale=0.8]{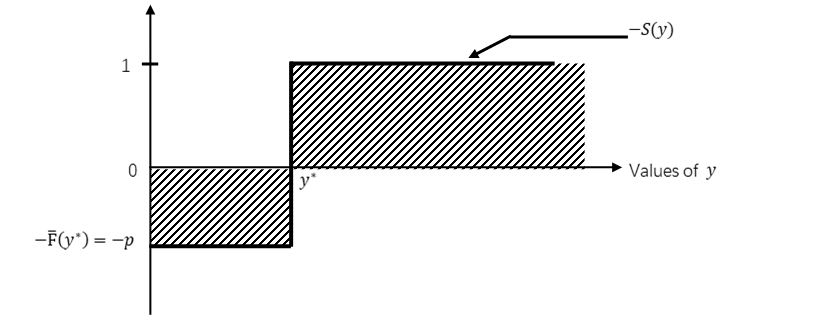}
  \caption{$\mathscr{M}$'s [$\mathscr{C}$'s] Payoff [Payback] Function $-S(y)$}
\end{figure}

Clearly, the payoff to $\mathscr{M}$ is a constant $+1$, irrespective of how much closer, to the required minimum $y^*$ the lifetimes are de Finetti's 2-sided bet therefore has the feature of providing an \textit{unjust} utility to $\mathscr{M}$, when it is invoked on lifetime's, and when $\mathscr{C}$ is a passive consumer.

The payoffs of Figures 1 and 2 are devoid of considerations pertaining to the costs of manufacture, the rewards of use, consequences of lost opportunities when the required minimum lifetime is not met, and other such economic matters. The focus of consideration  here is mainly the credibility of $\mathscr{M}$'s specified $\bar{F}(y)$. Indeed, Figures 1 and 2 provide an interpretation of the meaning of a lifetime distribution as seen from the perspective of $\mathscr{C}$ and $\mathscr{M}$, in terms of a de Finetti style 2-sided monetary bet. The two figures are also representative of $\mathscr{C}$ not being adversarial to $\mathscr{M}$ and $\mathscr{M}$ being amicable to $\mathscr{C}$. What would the payoff functions look like if $\mathscr{C}$ requires that the payoff to $\mathscr{M}$ encapsulate a better sense of being just, and $\mathscr{M}$ abides with this requirement? This matter is discussed next.

\subsection{Passive $\mathscr{C}$ and Amicable $\mathscr{M}$ with Just Payoffs}

A way to obviate the unjust feature of the payoffs given before is for $\mathscr{C}$ to make his(her) payoff function, no more the simple step-function of Figure 2. As will be seen later, this would also be the path for an adversarial $\mathscr{C}$ to express disagreement with $\mathscr{M}$'s specified $\bar{F}(y)$, $y\geq 0$. For example, suppose that the payoff to $\mathscr{C}$ is of the form indicated in Figure 3. This would correspond to a 180$\degree$ rotation of Figure 4, which is $\mathscr{M}$'s payoff function, when $\mathscr{C}$ is not adversarial to $\mathscr{M}$. The payoff function of Figure 4 reflects the feature of being more just (to $\mathscr{C}$) than that of Figure 2, because with the former, the larger the lifetime (over the minimum of $y^*$), the better the payoff to $\mathscr{M}$. Indeed, as shown in Figure 4, the payoff to $\mathscr{M}$ for lifetimes larger than $y^*$, is a concave increasing function of $y$ ($\geq y^*$). In the interest of simplicity, for values of $y<y^*$, the payoff (penalty) to $\mathscr{C}$($\mathscr{M}$) is assumed constant, but this too need not be so.
\newpage
\begin{figure}[!ht]
	\centering
	\includegraphics[scale=0.8]{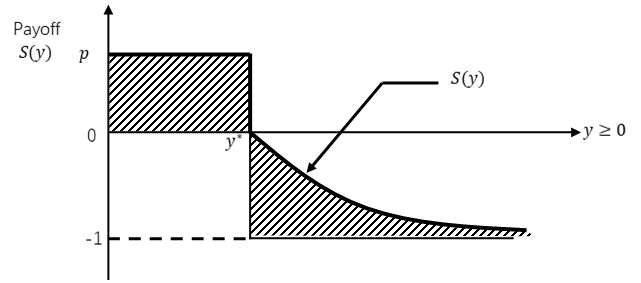}
	\caption{$\mathscr{C}$'s Payoff Function $S(y)$}
\end{figure}

\begin{figure}[!ht]
	\centering
	\includegraphics[scale=0.8]{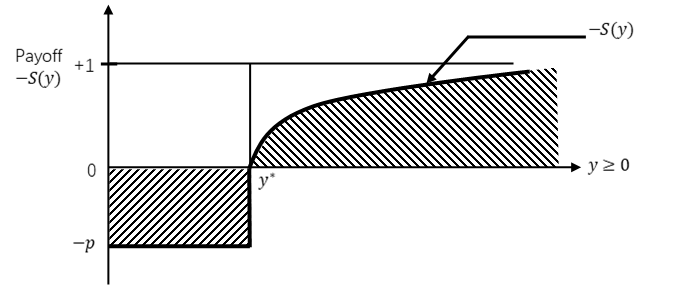}
	\caption{$\mathscr{M}$'s Payoff Function $-S(y)$}
\end{figure}

In Figures 3 and 4, the bounds $-1$ and $+1$ are arbitrary and so are the bounds $p$ and $-p$. Their purpose is to indicate a parallel with those of Figures 1 and 2, where, per de Finetti's set-up, the payoff functions are bounded by $-1$ and $+1$. As mentioned before, an inevitable consequence of any betting strategy based on money is the entanglement of probability and the utility for money. Thus, inherent to any choice by $\mathscr{C}$ of a payoff function $S(y)$, there is a probability distribution for $Y$, say $\bar{G}(y)$. In Section 4, we shall allude to the relationship between $\bar{G}(y)$ and $S(y)$. However, before doing so it may be useful to say a few words about considerations that may go in $\mathscr{C}$'s specification of an $S(y)$, $y\geq0$ -- $\mathscr{C}$'s state dependent utility function.

As is typical in reliability and survival analysis applications, it is common for $\mathscr{M}$ to first specify an $\bar{F}(y)$, and based on this, for $\mathscr{C}$ to accept or to reject $\mathscr{M}$'s offer. The uncertain entity in question is an item's lifetime for which $\mathscr{C}$ has specified a mission time $y^*>0$. Having specified $\bar{F}(y)$ and nothing more, an amicable $\mathscr{M}$'s disposition is to necessarily abide by the payoff function of Figure 2. If $\mathscr{C}$ is not adversarial to $\mathscr{M}$, then $\mathscr{C}$ abides by the payoff function of Figure 1, which is a 180$\degree$ rotation of Figure 2 around its horizontal axis. If in the interest of receiving a just payoff, $\mathscr{M}$ prefers to use the payoff function of Figure 4, and here again $\mathscr{C}$ chooses not to be adversarial to $\mathscr{M}$, then $\mathscr{C}$'s payoff function would be that of Figure 3, which is a rotation of Figure 4 around its horizontal axis. Similarly, were $\mathscr{C}$ to prefer the payoff function of Figure 3, and were $\mathscr{M}$ feel compelled to abide by $\mathscr{C}$'s choice, then $\mathscr{M}$'s payoff function would be a rotation of Figure 3 around its horizontal axis. The same is true of all other possible choices for $S(y)$, $y\geq0$.

To summarize, adversarial behavior between $\mathscr{C}$ and $\mathscr{M}$ is characterized here in terms of the payoff functions used. Whenever the payoff function of $\mathscr{C}$ is not a 180$\degree$ rotation of the payoff function of $\mathscr{M}$, and vice-versa, an adversarial scenario arises. Alternatively put, we see adversaries as those whose payoff functions are not \textit{rotationally symmetric}. Adversaries do not abide by what many would claim to be rules of fairplay.

\section{The Adversarial $\mathscr{C}$ and Amicable $\mathscr{M}$ Scenario}

Suppose that $\mathscr{C}$ has specified a $y^*>0$, and $\mathscr{M}$ has declared an $\bar{F}(y)$, $y\geq 0$. Based on these, and $\mathscr{M}$'s utility for money, $\mathscr{M}$ will arrive upon a payoff function $-S(y)$ of the forms illustrated in Figures 2 and 4. For purpose of discussion supposed that it is the ``just'' payoff of Figure 4 that appeals to $\mathscr{M}$. Suppose that $\mathscr{C}$ is adversarily dispositioned towards $\mathscr{M}$; then a payoff that is (a de Finetti style) symmetric rotation of Figure 4 will not be acceptable to $\mathscr{C}$. Instead, $\mathscr{C}$ will want to propose a payoff function that is more rewarding to $\mathscr{C}$ when $y<y^*$, and less punitive to $\mathscr{C}$ when $y\geq y^*$. The nature of what this payoff to $\mathscr{C}$ should be like, is the topic of this section. But first some words about the possible reasons underlying $\mathscr{C}$'s adversarial disposition.

First and formost, $\mathscr{C}$ may find $\mathscr{M}$'s $\bar{F}(y)$ overly optimistic and may thus want to challenge $\bar{F}(y)$ via a stress test. This would especially be so if $\mathscr{C}$ is a regulator who is mandated to thoroughly scrutinize $\bar{F}(y)$. $\mathscr{C}$ may also want to prove an $\mathscr{M}$ wrong with the intent of eliminating the $\mathscr{M}$. We propose that $\mathscr{C}$'s instrument for challenging $\mathscr{M}$ would be a ``stringent'' payoff function which penalizes $\mathscr{M}$ heavily when the observed $y<y^*$, and rewards $\mathscr{M}$ sparingly when $y\geq y^*$. Denote this payoff function by $S^*(y)$, and let $S^*(y)=A(y)S(y)$, where $A(y)>1$ when $y<y^*$ and $A(y)\leq 1$, when $y \geq y^*$; see Figure 5. We call $A(y)$ the \textit{adversarial stress function}. We illustrate, via Figure 6, $A(y)$'s effect on $\mathscr{M}$'s rotated payoff function $S(y)$, to produce $S^*(y)$ -- $\mathscr{C}$'s \textit{adversarial payoff function}. Observe that $A(y)$ exaggerates $S(y)$ for $y<y^*$, and dampens it for $y\geq y^*$. For reasons that will become clear in the sequel, we suppose that $A(y)\geq 0$, for all $y\geq 0$, and that $A(y)$ is also bounded above by $B>0$. Thus $0\leq A(y)\leq B \geq 1$. How best must $\mathscr{C}$ choose a meaningful $A(y)$ is the topic of Section 3.1. Figure 5 illustrates an archetypal form for $A(y)$, and Figure 6 its effect on $S(y)$ -- shown by the dashed lines of Figure 6.


\newpage
\begin{figure}[!ht]
	\centering
	\includegraphics[scale=0.8]{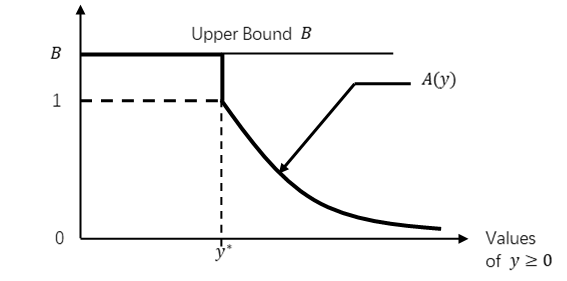}
	\caption{$\mathscr{C}$'s Adversarial Stress Function $A(y)$}
\end{figure}

\begin{figure}[!ht]
	\centering
	\includegraphics[scale=0.8]{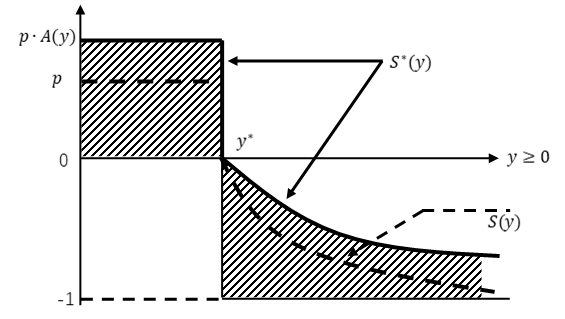}
	\caption{$\mathscr{C}$'s Adversarial Payoff Function $S^*(y)$}
\end{figure}

\subsection{The Adversarial Stress Function and Betting Scores}

The material which follows is an adaptation of some recent work by Shafer (2019) on statistical communication, adapted for the adversarial lifetime scenario considered here. It is based on the premise that when $\mathscr{C}$ and $\mathscr{M}$ agree to an exchange of goods -- henceforth the (adversarial) ``game'' -- $Y$ is not known and thus $A(Y)$ is a random quantity. Assume that $\mathscr{M}$ is confident of the survivability of his (her) product and is therefore amicable, but not naive, and agrees to $\mathscr{C}$'s modification of $S(y)$ to $S^*(y)$ as a payoff function. $\mathscr{M}$ then computes the expect value of $A(Y)$ with respect to $\mathscr{M}$'s $\bar{F}(y)$, namely $E_F[A(Y)]=\sum\limits_yA(y)f(y)$, and offers to sell the game to $\mathscr{C}$ at the price $E_F[A(Y)]$. If $\mathscr{C}$ accepts $\mathscr{M}$'s offer, then the amount risked by $\mathscr{C}$ is $E_F[A(Y)]$. Once the game is played, that is $Y$ observed as $y$, the quantity $A(y)/E_F[A(Y)]$ is $\mathscr{C}$'s \textit{adversarial betting score}. This is the factor by which $\mathscr{C}$'s amount risked gets multiplied. A large betting score can be seen as the best evidence $\mathscr{C}$ has against $\bar{F}(y)$; the larger the score, the stronger the evidence against $\bar{F}(y)$. $\mathscr{C}$'s aim therefore is to maximize the betting score, subject to $\mathscr{C}$'s limitation on the amount $\mathscr{C}$ is willing to risk. Since $\mathscr{C}$'s choice of $A(y)$ influences the betting score, $A(y)$ can be seen as the analogue of a physical stress in an accelerated stress test, or the level of dose in a dose-response test.

Since $A(y)$ is bounded by $B\geq 1$, $E_F[A(Y)]<\infty$, and thus $E_F[A(Y)]$ can be normalized to one. When such is the case, $\mathscr{C}$'s adversarial betting score is simply $A(y)$. To summarize, in the architecture describe here, there are three entities of concern to $\mathscr{C}$: an adversarial stress function $A(y)$, $y\geq 0$; an adversarial betting score $A(y)/E_F[A(Y)]$, and $S^*(y)-E_F[A(Y)]$, a payoff function adjusted for the cost to $\mathscr{C}$ of subjecting $\bar{F}(y)$ to a stress-test; note that $S^*(y)-E_F[A(Y)] \geq -1$. Whereas $\mathscr{M}$'s payoff is not of direct concern here, the -- $S(y)$ of Figure 4 should be increased by $E_F[A(Y)]$, this being an added reward to $\mathscr{M}$ for being amicable to $\mathscr{C}$'s stress test.

\subsection{Properties of Adversarial Stress Function: Choosing $A(y)$}

Let $q(y)=A(y)f(y)$, and note that $\sum\limits_y q(y) = \sum\limits_y A(y)f(y) = E_F[A(Y)] =1$. Furthermore, since $A(y)\geq 0$, it follows that $q(y)\geq 0$; this means that $q(y)$ can be seen as the probability mass function of some random variable, say $Z$. One can think of $Z$ as the random variable that would have directly yielded $\mathscr{C}$, the adversarial payoff $S^*(y)$. Since $A(y)$ is also $\mathscr{C}$'s betting score -- when $E_F[A(y)]$ is normalized to one, one can also see $A(y)$ as the \textit{likelihood ratio} of the probability mass functions of the random variable $Z$ and the random variable $Y$. Bear in mind that the probability mass of $Y$ is specified by $\mathscr{M}$, whereas that of $Z$ is specified by $\mathscr{C}$. An equivalence between a likelihood ratio and a betting score could be a noteworthy observation.

Since the betting score $A(y)$ is the factor by which the money risked by $\mathscr{C}$ gets multiplied, $A(y)$ spawns a utility to $\mathscr{C}$, say $U[A(y)]$. The bigger the $A(y)$ the bigger the utility. Suppose that this utility is logarithmic, namely, $U[A(y)]=\log_2A(y)$. This utility encapsulates the satisfaction that $\mathscr{C}$ derives in outfoxing $\mathscr{M}$, or in justifying $\mathscr{C}$ as an effective regulator. The utility function need not be logarithmic, but assuming so leads to information theoretic considerations; these are articulated below.

In decision theory one aims to make choices that maximize an expected utility, namely, the expected value of $\log_2A(y) = log_2 (q/f)$. However, by the converse of \textit{Gibbs Inequality} [cf. Kullback (1959)], for any utility like $log_2 (q/f)$, taking an expectation with respect to anything other than $q$ will not maximize the expected utility. We are thus motivated to maximize $E_q[log_2\frac{q}{f}]=D_{K-L}(q:f)$, the \textit{Kullback-Leibler discrimination} between $q$ (which is a probability), and $f$ (taken as reference) [see Kullback and Leibler (1951)]. Thus if $\mathscr{C}$ wishes to subject $f$ to the most severe adversarial stress-test as is possible, then $\mathscr{C}$'s $A(y)$ should be such that $D_{K-L}(q:f)$ is maximized. Since $q(y)=A(y)f(y)$, $D_{K-L}(q:f)=E_q[\log_2A(y)]$, $\mathscr{C}$'s aim would be to choose that $A(y)$ which maximizes $\sum\limits_y\log_2A(y)q(y)$, or equivalently minimizes $-\sum\limits_y\log_2A(y)q(y)$, which is like the cross-entropy of $A(y)$, the adversarial stress function, with respect to the probability $q(y)$.

As an illustration of the workings of the above, suppose that $A(y)=1$, for all $y\geq0$. Now, $q=f$, and $E_q[\log_2 1]=0$; this means that $\mathscr{C}$ will garner a non-zero betting score only when $q\neq f$. When $A(y)=0$, for all $y\geq 0$, the maximum expected utility is $-\infty$; indeed this is so, if for any $y$, $A(y)=0$. Thus we require that $A(y)>0$ for all $y\geq0$. Finally, for any $B>0$, the maximum expected utility assuming $A(y)=B$, for all $y$, will be $\log_2 B$, suggesting that the larger the upper bound on $A(y)$, the larger the expected utility. But large values of $B$ increase the amount risked by $\mathscr{C}$, namely $E_F[A(Y)]$. This in turn places a restriction on how much larger than one $B$ can be. Once an $A(y)$ is pinned down by $\mathscr{C}$, with $0<A(y)\leq B>1$, the cost adjusted payoff to $\mathscr{C}$, $S^*(y)-E_F[A(Y)]$ gets defined, and the game gets played. By this we mean that $Y$ gets observed as $y$, and based on what $y$ is, $\mathscr{C}$ makes a choice as to whether to certify or not $\mathscr{M}$'s specified $\bar{F}(y)$. Some strategies for $\mathscr{C}$'s operationalize of adversarial stress test are given below.

\section{Operationalizing the Adversarial Stress Test} 

Since $\mathscr{C}$'s cost adjusted payoff is $S^*(y)-E_F[A(Y)] = A(y)S(y)-E_F[A(Y)]$, we could suppose that when $y<y^*$, $\mathscr{C}$ should not bear the burden of having to pay $\mathscr{M}$ the amount $E_F[A(Y)]$ for the stress test. This of course presumes that both $\mathscr{M}$ and $\mathscr{C}$ have agreed to the stress test. However, for the case $y\geq y^*$, $\mathscr{C}$ should subsidize the cost of the stress test as a way to compensate $\mathscr{M}$ for $\mathscr{C}$'s adversarial and distrustful disposition. Thus $\mathscr{C}$'s risk adjusted payoff function would take the form shown in Figure 7; compare Figure 6 to Figure 7.
\newpage
\begin{figure}[!ht]
  \centering
  \includegraphics[scale=0.8]{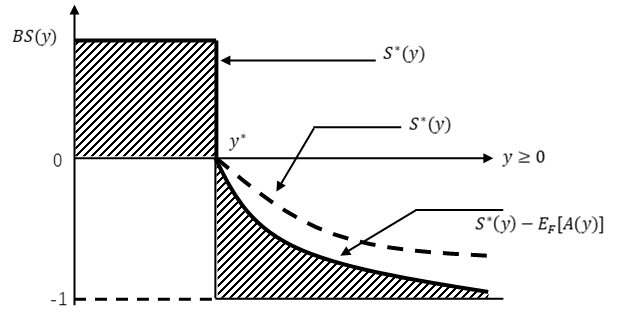}
  \caption{$\mathscr{C}$'s Risk Adjusted Payoff Function}
\end{figure}

There can be several strategies for $\mathscr{C}$ to operationalize the import of Figure 7, vis a vis certifying or not $\mathscr{M}$'s $\bar{F}(y)$. Clearly, if $y<y^*$, $\mathscr{C}$ will be reluctant to certify $\bar{F}(y)$. When $y\geq y^*$, $\mathscr{C}$ is faced with a challenge, namely, for what values of $y$, $y\geq \tilde{y} \geq y^*$ should $\bar{F}(y)$ be certified? The answer could depend $\mathscr{C}$'s choice of the risk adjusted payoff function. The larger the $y$, the larger is $S^*(y)-E_F[A(Y)]$, which means the larger the amount that $\mathscr{C}$ has to payback $\mathscr{M}$. This would suggest that $\tilde{y}$ should be as close to $y^*$ as is meaningful. Should $\mathscr{C}$ want to limit the payback to $\mathscr{M}$ at some $C>-1$, then $\tilde{y}$ would be that $y$ for which $S^*(\tilde{y})-E_F[A(Y)]=C$. Consequently, $\mathscr{C}$ would certify $\bar{F}(y)$ whenever the observed lifetime $y \geq \tilde{y}$.

The strategy proposed above does not take into consideration $\bar{G}(y)$, $\mathscr{C}$'s survival function of the item in question. Assuming that $\mathscr{C}$ has in mind a $\bar{G}(y)$, it makes sense to assume that since $\mathscr{C}$ is adversarial to $\mathscr{M}$, $\bar{F}(y) \geq \bar{G}(y)$, for $y\geq 0$; otherwise, it does not make sense to stress test. This in turn would suggest that $\bar{G}(y^*) \leq \bar{F}(y^*)$, so that certifying $\bar{F}(y)$ when the observed $y\geq y^*$ will be more optimistic than what $\mathscr{C}$ believes the lifetimes of the item will be. One possibility is for $\mathscr{C}$ to pin down that $\tilde{y}$ for which $\bar{G}(\tilde{y}) = \bar{F}(y^*)$, and certify $\bar{F}(y)$ if the observed $y\geq \tilde{y} \geq y^*$. This schemata will also enable $\mathscr{C}$ to certify $\bar{F}(y)$ with a high degree of certitude (confidence), by considering several stress tests, say n, and requiring that for at least $k\leq n$ of these tests, the observed $y \geq \tilde{y}$; here $\mathscr{C}$'s probability that $y\geq \tilde{y}$ is $\bar{G}(\tilde{y})$, and $n$ and $k$ be chosen to assure a specified degree of certitude.

\section*{\large Acknowledgements}

Supported by a grant from the City University of Hong Kong, Project Number 9380068 and the Theme Based Research Scheme Grant T 32-102/14N and T 32-101/15R. Professor Glen Shafer of Rutgers University provided a pre-print of his paper, which motivated and stimulated the work discribed here. I thank him for this gesture.

\newpage
\section*{\large References}
\quad \\ [-100pt]

\end{document}